\documentclass{article} 
\usepackage{iclr2021_conference,times}

\usepackage{amsmath,amsfonts,bm}

\def\eqref#1{equation~\ref{#1}}

\def\1{\bm{1}}

\def\vc{{\bm{c}}}

\def\vt{{\bm{t}}}

\def\vx{{\bm{x}}}

\DeclareMathAlphabet{\mathsfit}{\encodingdefault}{\sfdefault}{m}{sl}
\SetMathAlphabet{\mathsfit}{bold}{\encodingdefault}{\sfdefault}{bx}{n}

\usepackage{hyperref}
\usepackage{url}
\usepackage{graphicx}
\usepackage{longtable}
\usepackage{booktabs}
\usepackage{placeins}
\usepackage{svg}
\usepackage{multirow}

\usepackage[group-separator={,}]{siunitx}

\title{Failing Conceptually: Concept-Based Explanations of Dataset Shift}

\author{Maleakhi A. Wijaya, Dmitry Kazhdan, Botty Dimanov \& Mateja Jamnik \\
Department of Computer Science and Technology\\
University of Cambridge\\
Cambridge, CB3 0FD, UK \\
\texttt{\{maw219,dk525,btd26\}@cam.ac.uk, mateja.jamnik@cl.cam.ac.uk}}

\iclrfinalcopy 
\begin{document}

\maketitle

\begin{abstract}
Despite their remarkable performance on a wide range of visual tasks, machine learning technologies often succumb to data distribution shifts. Consequently, a range of recent work explores techniques for detecting these shifts. Unfortunately, current techniques offer no explanations about what triggers the detection of shifts, thus limiting their utility to provide actionable insights. In this work, we present \textbf{C}oncept \textbf{B}ottleneck \textbf{S}hift \textbf{D}etection (CBSD): a novel explainable shift detection method. CBSD provides explanations by identifying and ranking the degree to which high-level human-understandable concepts are affected by shifts. Using two case studies (\textit{dSprites} and \textit{3dshapes}), we demonstrate how CBSD can accurately detect underlying concepts that are affected by shifts and achieve higher detection accuracy compared to state-of-the-art shift detection methods.
\end{abstract}

\section{Introduction}

Despite their remarkable performance on a wide range of visual tasks~\citep{liu2017survey}, machine learning technologies are often hampered by  \textit{data distribution shifts}~\citep{rabanser2018failing}. Particularly, when faced with changes between the \textit{source} (distribution from which training data is sampled) and \textit{target} (distribution from which real-world data is sampled) data distributions, state-of-the-art models often have their performance compromised, as demonstrated by \citet{szegedy2013intriguing}. The dataset shifts are recurrent in predictive modelling and lead to serious consequences if left unresolved (e.g., deployment of uncertain models that fail in critical domain applications~\citep{zech2018variable,schulam2017reliable}). 

Various approaches have been developed to detect dataset shifts including: statistical-based approaches \citep{quionero2009dataset}, which calculate the statistical distance between source and target distributions; \textit{domain classifiers} \citep{quionero2009dataset,rabanser2018failing}, which explicitly train classifiers to distinguish instances from source and target distributions; and \textit{anomaly detection}, which identifies whether a test instance significantly differs from the majority of the source data (see \citet{chandola2009anomaly,markou2003novelty,ruff2020unifying} for surveys and \citet{breunig2000lof,liu2008isolation,scholkopf1999support} for examples).
Although numerous dataset shift detection methods have been proposed, none of them provide any explanations about data characteristics that lead to the shift. Shift explanations could be valuable in many cases; for instance, identifying the human-understandable concepts~\citep{ghorbani2019towards} that are affected by shifts is useful for inferring the type of shifts, determining whether the shifts are malicious, and guiding the data gathering procedure (see Figure \ref{fig:pipeline} for an example).

\textbf{Contributions} 
We propose a novel method for detecting and explaining dataset shifts on visual tasks, called \textbf{C}oncept \textbf{B}ottleneck \textbf{S}hift \textbf{D}etection (CBSD)\footnote{\url{https://github.com/maleakhiw/explaining-dataset-shifts}}, which comprises of three main contributions: (1) to the best of our knowledge, we are the first to present an explainable shift detection method; (2) our method achieves higher detection accuracy than the state-of-the-art black-box shift detection methods (BBSD) \citep{lipton2018detecting,rabanser2018failing}; (3) in contrast to BBSD, our method can detect a new type of shift based on factors of variation that may not be correlated to the end task.  We empirically demonstrate the capabilities of our method and the limitations of BBSD using two datasets (\textit{dSprites} and \textit{3dshapes}).


\section{Methodology}

\begin{figure}[t!]
\begin{center}
\includegraphics[width=1\linewidth]{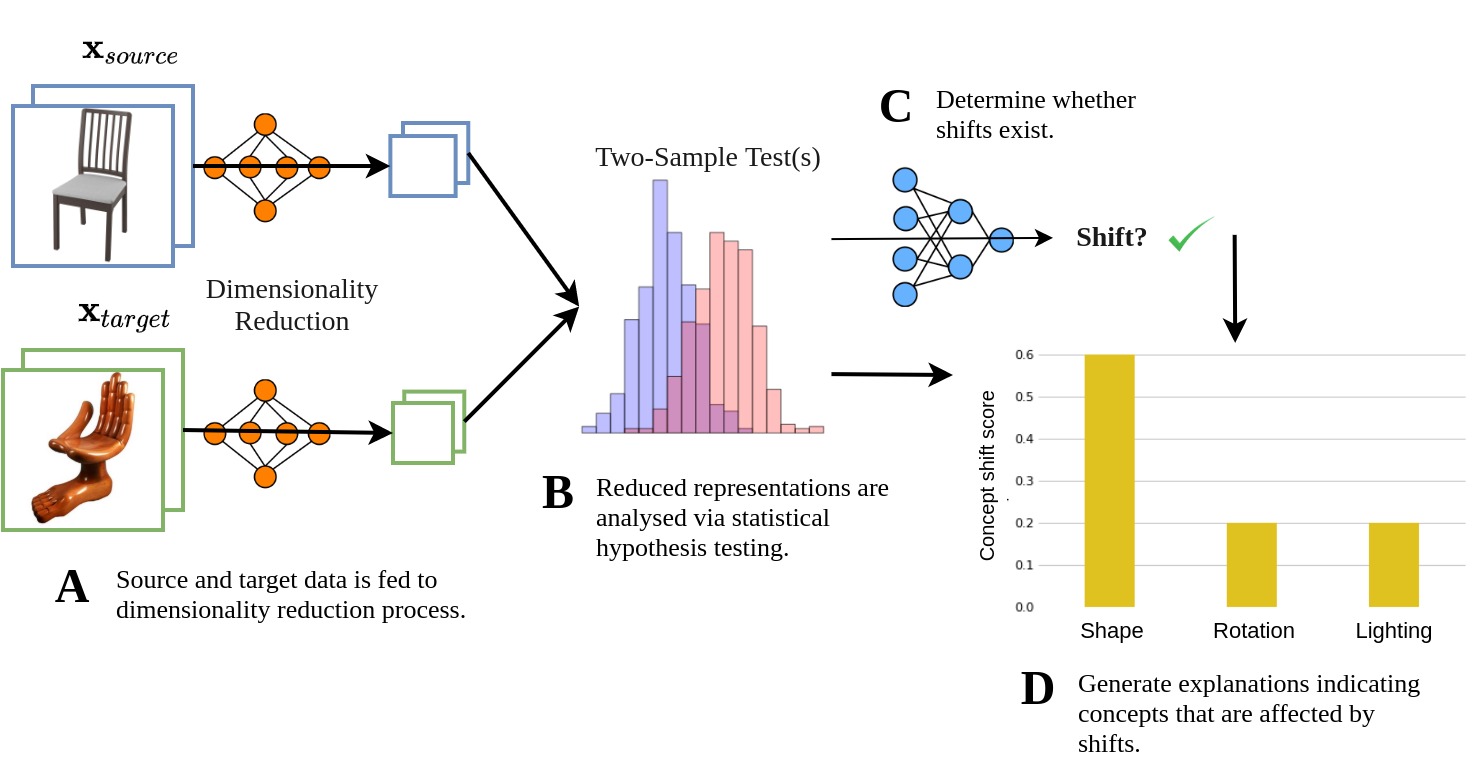}
\end{center}
\caption{Shift detection pipeline. Let us assume the source data consists of images of ordinary chairs, whilst target data consists of exotic, unconventional chairs. A: The source and target data are fed to a dimensionality reduction process. B: The reduced representations are analysed using two-sample hypothesis testing, producing $p$-value and test statistics. C: The resulting $p$-value and test statistics are used to determine whether shifts exist. D: CBSD-based methods provide explanations for detected shifts, by listing concepts most affected by the shift (in this case - the ``shape" concept).}
\label{fig:pipeline}
\end{figure}

Given labelled data from the source distribution $\{(\vx_1, y_1), \dots (\vx_n, y_n)\} \sim p$ and unlabelled data from the target distribution $\{\vx^\prime_1, \dots \vx^\prime_m\} \sim q$, we aim to test the equivalence of the source distribution $p$ and the target distribution $q$. Formally, $\mathbf{H}_0: p(\vx) = q(\vx^\prime)$ vs $\mathbf{H}_1: p(\vx) \neq q(\vx^\prime)$. Figure~\ref{fig:pipeline} illustrates our shift detection pipeline.

Motivated by recent successes of BBSD \citep{lipton2018detecting}, \citet{rabanser2018failing} proposed to use the outputs of a deep neural network (DNN) classifier trained on the source data as the dimensionality-reduced representation. Two versions of the method exist: \textit{BBSDs}, which uses the softmax outputs and \textit{BBSDh}, which uses the hard-thresholded prediction. The overall procedure involves feeding the source (target) data to the trained classifier, where its outputs are used as reduced representations for the source (target) for subsequent hypothesis testing. In the study, \citet{rabanser2018failing} demonstrated that the method attained the best shift detection accuracy than other methods.

Recent work on concept-based explanations~\citep{kazhdan2020now,koh2020concept} proposed the idea of \textit{concept decomposition}, in which input data is mapped into an interpretable concept representation. \citet{kazhdan2020now} presented concept model extraction (CME), decomposing a pretrained DNN $f: \mathcal{X} \rightarrow \mathcal{Y}$, $(\mathcal{X} \subset \mathbb{R}^n, \mathcal{Y} \subset \mathbb{R}^o)$, into a composition of functions $g: \mathcal{X} \rightarrow \mathcal{C}$, $(\mathcal{C} \subset \mathbb{R}^k)$, mapping input data to their concept representation and $h: \mathcal{C} \rightarrow \mathcal{Y}$, mapping concept representation to output space, such that $f(\vx) = h(g(\vx))$. Instead of approximating existing models, \citet{koh2020concept} suggested the concept bottleneck model (CBM): an inherently interpretable model build from scratch comprising two sub-models, where the first predicts an intermediate set of human-specified concepts $\vc \in \mathcal{C}$ and the second uses $\vc$ to predict the final output $y \in \mathcal{Y}$. \citet{koh2020concept} showed that CBMs attained comparable performance with end-to-end models.


In this work, we propose the CBSD, which builds on top of BBSD, but relies on concept representations as the dimensionality reduced representation. For dimensionality reduction, we use concept extraction methods~\citep{kazhdan2020now, koh2020concept} with the added benefit of explainability (since each dimension is interpretable). We introduce two version of CBSD: \textit{CBSDs} which uses the softmax outputs, and \textit{CBSDh}, which uses the hard-thresholded prediction of concepts.

We apply statistical tests for each source and target concept representations to determine if a distribution shift exists between them. We identify the concepts that are affected by shifts using the $p$-value and the extent to which they experience the shift using the novel \textit{concept shift score} (CSS). Let $k$ be the number of distinct concepts. Given the test statistics for all concepts $\vt$, the CSS for the $i$th concept is expressed as:
    
    \begin{equation}
        \textit{CSS}(\vt, i) = \frac{t_i}{\sum_{l=1}^{k} t_l}
    \end{equation}
    
A higher CSS indicates that the concept is more affected by the shifts relative to others. CBSD determines that dataset shift exists if at least one of the concepts is shifted.

\section{Experiments}

\textbf{Datasets \& Tasks} We conducted experiments on dSprites~\citep{dsprites17} and 3dshapes~\citep{3dshapes18} to compare the shift detection accuracy of CBSD with other methods and to verify that it can accurately detect and rank concepts affected by shifts. For dSprites, we rely on the task setup from \citet{kazhdan2020now}, which predicts the ``shape" concept. For 3dshapes we define the task for discriminating between all possible ``scale" and ``shape" concept value combinations. Further details are given in Appendix~\ref{appendix:exp}. 

\textbf{Benchmarks} We compared CBSD to the following state-of-the-art dimensionality reduction (DR) benchmarks from \cite{rabanser2018failing}: principal component analysis (PCA), sparse random projection (SRP), BBSDs and BBSDh. Similarly to \cite{rabanser2018failing}, we used different statistical tests, namely maximum mean discrepancy \citep{gretton2012kernel} and Kolmogorov-Smirnov test + Bonferroni correction \citep{weisstein2004bonferroni} for evaluating multidimensional representations (PCA, SRP, BBSDs, and CBSDs), and Chi-squared test for evaluating unidimensional representations (BBSDh and CBSDh). 

\textbf{Shifts} We simulated various shifts defined by \citet{rabanser2018failing}, including Gaussian, knockout, image, and adversarial shifts. Additionally, we introduced a novel type of shift, referred to as \textit{concept shift}, which represents realistic real-world shifts, attacking specific concepts in images. We evaluate various methods' abilities to detect shifts at $5\%$ significance level ($\alpha=0.05$). The datasets were randomly split into training, validation, and test sets. The training data was used to fit the DR methods. Shift detection was performed on the reduced representations of the validation and test sets. The shifts were only applied to the test data - the test data was treated as real-world data, which may or may not have been subject to shifts. We included the no-shift case to check against false positives. Further details regarding the experimental setup can be found in Appendix~\ref{appendix:exp}.

\section{Results and Discussion}
\label{sec:result_discussion}
\begin{figure}[t!]
\setlength\tabcolsep{0pt}
\begin{tabular}{ccc}
    \multicolumn{3}{c}{\includegraphics[width=.65\linewidth]{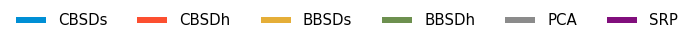}} \\
  \includegraphics[width=.33\linewidth]{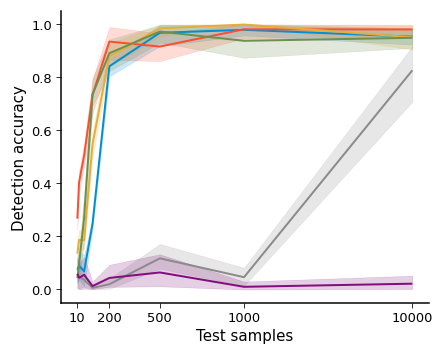} &   \includegraphics[width=.33\linewidth]{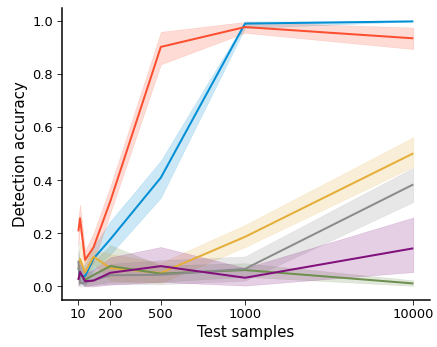} & \includegraphics[width=.33\linewidth]{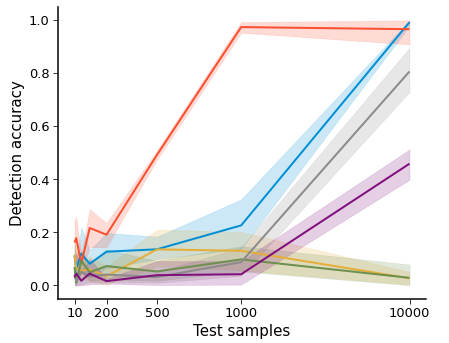} \\
{\scriptsize (a) Knockout shift\par} & {\scriptsize (b) Concept (scale) shift \par} & {\scriptsize (c) Concept (x, y) shift  \par} \\
\end{tabular}
\caption{Shift detection accuracy for \textbf{medium} knockout, concept (scale), and concept (x, y) shifts. Plots show mean values over 100 random runs with a 95\% confidence interval error bar.}
\label{fig:acc}
\end{figure}

Firstly, we evaluate the detection accuracy of CBSD, and compare it to other benchmarks. The shift detection accuracy was expressed as the number of times that shift detection algorithms correctly classify where a shift exists divided by the total number of experiments~\citep{rabanser2018failing}. Overall, detection accuracy increases with larger shift intensities and test sample sizes (Figure~\ref{fig:acc}). 

Importantly, we observe that CBSD methods successfully detected shifts targeted at concept covariates (Figure~\ref{fig:acc} (b) and (c)), whilst BBSD methods could not detect them irrespective of the number of test samples (BBSDh could not detect these at all, whilst BBSDs could detect these only in some cases, given large test set samples). Furthermore, CBSD methods successfully detected shifts targeted at task labels, whilst PCA or SRP methods failed (Figure \ref{fig:acc} (a)). 


\begin{figure}[t!]
\setlength\tabcolsep{0pt}
\begin{tabular}{ccc}
    \multicolumn{3}{c}{\includegraphics[width=.25\linewidth]{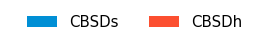}} \\
  \includegraphics[width=.33\linewidth]{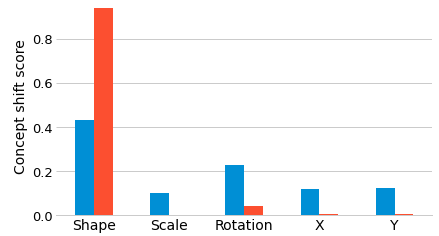} &   \includegraphics[width=.33\linewidth]{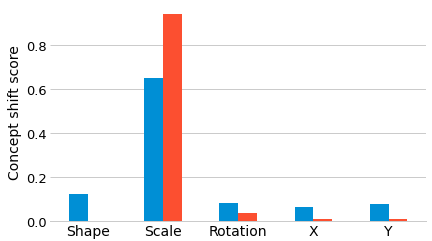} & \includegraphics[width=.33\linewidth]{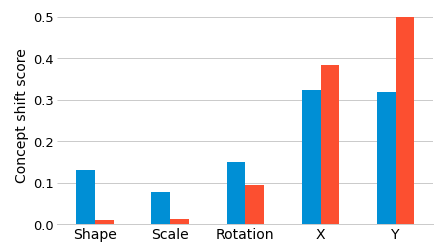} \\
{\scriptsize (a) Knockout shift\par} & {\scriptsize (b) Concept (scale) shift \par} & {\scriptsize (c) Image (translation) shift  \par} \\
\end{tabular}
\caption{Concept shift scores for knockout, concept (scale), and image (translation) shifts. The CBSDs and CBSDh accurately identified and ranked concepts that were affected by shifts.}
\label{fig:explanation-barplot}
\end{figure}

Secondly, we evaluate the ability of CBSD methods to provide meaningful shift detection explanations.  Figure~\ref{fig:explanation-barplot} shows the CSS for each concept, calculated using CBSDh and CBSDs when target data was subjected to knockout, concept (scale), and image (translation) shifts. Overall, we observe that both CBSDh and CBSDs accurately identified concepts that were most affected by the shift: they recovered that ``shape" was the most affected concept when the target data was subjected to a knockout shift; ``scale" was the most affected concept when the target data was subjected to a concept (scale) shift; ``x" and ``y" were the most affected concepts when the target data was subjected to an image (translation) shift.  Further results are shown in Appendix \ref{appendix:result}.


Overall, these results demonstrate that BBSD methods work particularly well for shifts that directly affect the task labels (Figure~\ref{fig:acc}~(a)). In cases where shifts did not directly affect task labels, their performance declined, with PCA or SRP approaches showing superior performance instead (Figure~\ref{fig:acc}~(c)). This can be explained by the fact that end-to-end DNN classifiers used by BBSD are likely to only retain the bare minimum information necessary to achieve a high predictive accuracy, and will not include any other information. Conversely, standard DR methods, such as PCA or SRP are not trained for a particular prediction task, enabling it to detect any type of shifts, although at a lower accuracy than BBSD for shifts that directly affect the task labels. The concept-based classifiers bridge the advantages of BBSD and standard DR methods; they store information pertaining to underlying dataset concepts that is both useful for prediction and sufficiently capture data covariates. Consequently, they were able to achieve a high detection accuracy for both types of shifts.

\section{Conclusion}

We presented CBSD, a novel explainable shift detection method capable of providing explanations by identifying and ranking the degree to which human-understandable concepts are affected by shifts. The resulting explanations are useful for inferring the shift types and guiding future data collection procedure, as they inform data collectors the concepts that they should focus on when collecting data. We discussed the strengths and limitations of state-of-the-art BBSD methods and showed how CBSD can be used to overcome their limitations, while preserving their performance. We demonstrated that CBSD methods attained similar or higher detection accuracy against BBSD and standard DR methods in all cases, whilst also providing explanations for the most affected underlying concepts. 

\bibliography{iclr2021_conference}
\bibliographystyle{iclr2021_conference}

\appendix

\section{Experimental Setup}
\label{appendix:exp}

\begin{figure}[h!]
\setlength\tabcolsep{0pt}
\centering
\begin{tabular}{cc}
      \includegraphics[width=.33\linewidth]{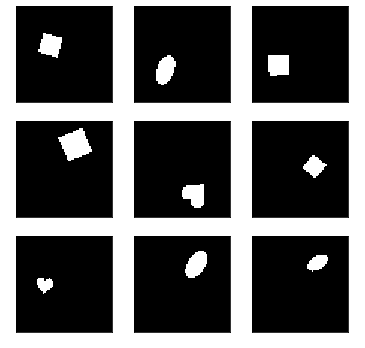} &   \includegraphics[width=.33\linewidth]{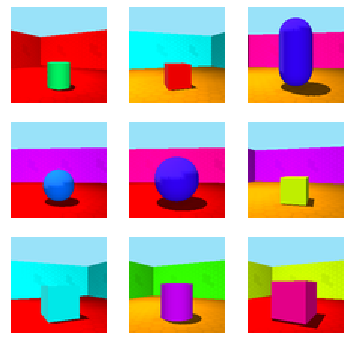} \\
    {\scriptsize (a) dSprites \par} & {\scriptsize (b) 3dshapes \par} \\[6pt]
     \includegraphics[width=.33\linewidth]{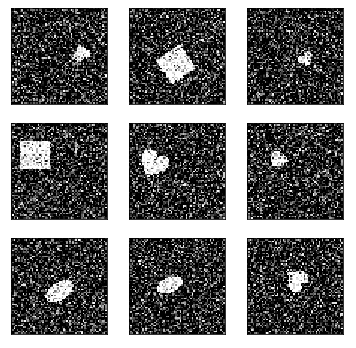} &   \includegraphics[width=.33\linewidth]{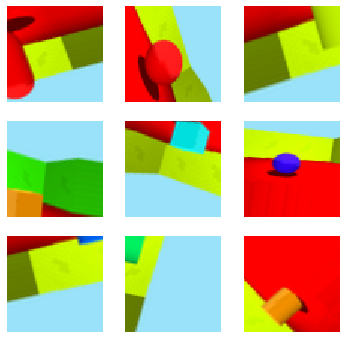} \\
    {\scriptsize (c) dSprites (Gaussian shift)  \par} & {\scriptsize (d) 3dshapes (image shift)  \par} \\[6pt]
    \end{tabular}
\caption{Sample dSprites and 3dshapes images. Subfigures (a), (b) show original images; subfigures (c), (d) show shifted images.}
\label{fig:dataset-img}
\end{figure}

\textbf{Datasets} The dSprites \citep{dsprites17} dataset comprises of 2D $64 \times 64$ pixel black-and-white shape images, generated from all possible combinations of six latent factors -- color, shape, scale, rotation, x, and y positions (Figure \ref{fig:dataset-img} (a)). Experiments were conducted on a subset of the dataset, comprising of $\num{100000}$ instances. We randomly partitioned the data into training ($\num{70000}$), validation ($\num{15000}$), and test ($\num{15000}$) splits. We define a classification task to determine the shape given an input image (see task 1~\citep{kazhdan2020now}). For each image sample, we define its task label as the shape concept label, which can be square, ellipse, or heart. The latent factors were chosen as concept annotations. Table \ref{fig:dsprites} shows all possible concepts and values. Note that the color concept was removed, as it is constant.

\begin{table}[h!]
    \centering
    \caption{dSprites concepts and values}
    \label{fig:dsprites}
    \begin{tabular}{cc}
    \\
    \toprule
     \textbf{Concept name} & \textbf{Values} \\ 
     \bottomrule
     Color & white \\
     Shape & square, ellipse, heart \\
     Scale & 6 values linearly spaced in $[0.5, 1]$  \\
     Rotation & 40 values in $[0, 2\pi]$ \\
     Position X & 32 values in $[0, 1]$ \\
     Position Y & 32 values in $[0, 1]$ \\
     \bottomrule \\
    \end{tabular}
\end{table}

The 3dshapes \citep{3dshapes18} dataset comprises of $64 \times 64$ pixel coloured 3D shape images, generated from all possible combinations of six latent factors -- floor hue, wall hue, object hue, scale, shape, and orientation (Figure~\ref{fig:dataset-img} (b)). Experiments were conducted on a subset of the dataset, comprising $\num{50000}$ instances. We randomly partitioned the data into training ($\num{30000}$), validation ($\num{10000}$), and test ($\num{10000}$) splits. We define a classification task for discriminating between all possible scale and shape concept value combinations. For each image sample, we define its task label as the scale and shape value combinations, which can be one of 32 values. The task is chosen such that it closely parallels experiments (task 2) defined in \citet{kazhdan2020now}. The latent factors were chosen as concept annotations. Table~\ref{tbl:3dshapes} shows all possible concepts and values.

\begin{table}[h!]
    \centering
    \caption{3dshapes concepts and values}
    \label{tbl:3dshapes}
    \begin{tabular}{cc}
    \\
    \toprule
     \textbf{Concept name} & \textbf{Values} \\ 
     \bottomrule
     Floor hue & 10 values linearly spaced in $[0, 1]$ \\
     Wall hue & 10 values linearly spaced in $[0, 1]$ \\
     Object hue & 10 values linearly spaced in $[0, 1]$  \\
     Scale & 8 values linearly spaced in $[0, 1]$ \\
     Shape & 4 values in $[0, 1, 2, 3]$ \\
     Orientation & 15 values linearly spaced in $[-30, 30]$ \\
     \bottomrule \\
    \end{tabular}
\end{table}

\FloatBarrier
\textbf{Dimensionality Reductors} We trained a convolutional neural network model \citep{le1989handwritten} for both tasks. Both models had the same architecture, consisting of 3 convolutional layers with $64-8 \times 8$, $128-6 \times 6$, and $128-5 \times 5$ filters, followed by 2 fully connected layers with ReLUs, and a softmax output layer. We applied $50\%$ dropout \citep{srivastava2014dropout} and batch normalisation. The models were trained for 200 epochs with early stopping. They achieved $100\%$ accuracy on their respective held-out evaluations for the original tasks respectively. The aforementioned models were used as the dimensionality reductors for both BBSDs and BBSDh.

For CBSD, we trained \textit{sequential} multi-task CBM models, in which a concept predictor is first trained to predict the provided concepts, and a label predictor is then trained to predict the task labels from the concept predictions. The input-to-concept models have the same architecture as the BBSD models, except for the softmax layers, where we provided separate output layers for predicting the concepts. The output-to-concept model is a logistic regression that takes concept values as inputs to predict the task label. Tables \ref{tab:concept_accuracy} show the concept accuracies for the CBM models. They achieved $100\%$ and $91\%$ accuracies on their respective held-out evaluations for the original tasks respectively. The input-to-concept models were used as the dimensionality reductors for both CBSDs and CBSDh.

For PCA and SRP, we reduced the dimensionality to 42 (for dSprites) and 8 (for 3dshapes), which explains roughly $80\%$ of total variance for their respective datasets.

\begin{table}[h!]
    \centering
    \caption{The CBMs held-out concept accuracies for dSprites and 3dshapes datasets.}
    \label{tab:concept_accuracy}
    \begin{tabular}{ccc}
    \\
    \toprule
    \textbf{Dataset} & \textbf{Concept name} & \textbf{Accuracy} \\ 
     \bottomrule
     
     \multirow{6}{*}{dSprites} & Color & $100\%$ \\
     &Shape & $100\%$ \\
     &Scale & $100\%$  \\
     &Rotation & $52\%$ \\
     &Position X & $87\%$ \\
     &Position Y & $88\%$ \\ \midrule
     
    \multirow{6}{*}{3dshapes} & Floor hue & $100\%$ \\
     &Wall hue & $100\%$ \\
     &Object hue & $100\%$  \\
     &Scale & $100\%$ \\
     &Shape & $100\%$ \\
     &Orientation & $100\%$ \\
     \bottomrule \\
    \end{tabular}
\end{table}

\FloatBarrier
\textbf{Experiment Configurations} We simulated various shifts with different intensities, affecting a fraction $\delta$ of test samples. The experiment setup closely follows the configurations discussed in \citet{rabanser2018failing}, except with the addition of concept shift, which we proposed. Details on the configurations are as follows: 

\begin{itemize}
    \item The number of test samples: 10, 20, 50, 100, 200, 500, 1000, 10,000.
    \item The shift intensities: small, medium, and large.
    \item The shift proportions $(\delta)$ of test data affected by shifts: $10\%, 50\%$, and $100\%$.
    \item The shift types:
    \begin{itemize}
        \item Gaussian noise: we corrupted a fraction $\delta$ of image covariates of the test sets by Gaussian noise with a standard deviation $\sigma \in \{1, 10, 100\}$ depending on the shift intensities.
        \item Knockout: we removed a fraction $\delta$ of test samples from a particular class, creating class imbalance. By default, we removed instances of the majority class.
        \item Concept: we added or removed a fraction $\delta$ of test set samples from a particular concept value, creating concept imbalance. For example, to shift the color concept, we may remove all \textit{red} object instances, creating a class imbalance in this concept. The concept shifts are created using domain knowledge, for example, we may introduce ``chair" concept shift upon inspecting that there are various significantly different types of chairs in a manufacturers' catalogues.
        \item Image: we explored more natural shifts to images, including translation, rotation, zoom, shear, and flip. In the experiments, we tested the shifts in isolation and combination with other shifts. The amount of translation, rotation, zoom, and shear are controlled by shift intensities.
        \item Combination: we considered combinations of different shift types defined above.
    \end{itemize}
\end{itemize}

We ran each configuration 100 times and averaged the results. Subsequently, the process is repeated five times to obtain the standard deviation and $95\%$ confidence interval. 

\section{Additional Results}
\label{appendix:result}

\subsection{dSprites}
\label{app:dsprites}

\begin{figure}[h!]
\setlength\tabcolsep{0pt}
\begin{tabular}{ccc}
    \multicolumn{3}{c}{\includegraphics[width=.25\linewidth]{figures/legend-bar.png}} \\
  \includegraphics[width=.33\linewidth]{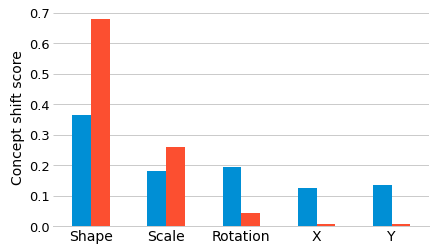} &   \includegraphics[width=.33\linewidth]{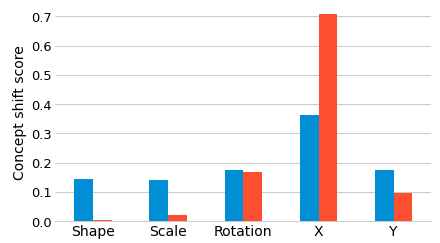} & \includegraphics[width=.33\linewidth]{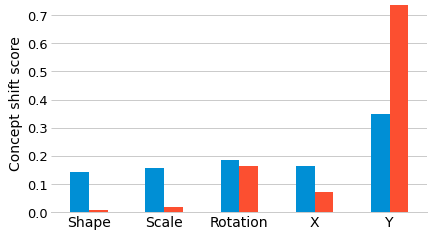} \\
{\scriptsize (a) Concept (shape, scale) shift\par} & {\scriptsize (b) Concept (x) shift \par} & {\scriptsize (c) Concept (y) shift  \par} \\[6pt]
 \includegraphics[width=.33\linewidth]{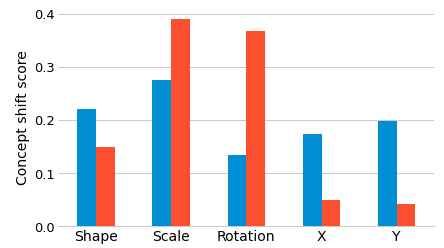} &   \includegraphics[width=.33\linewidth]{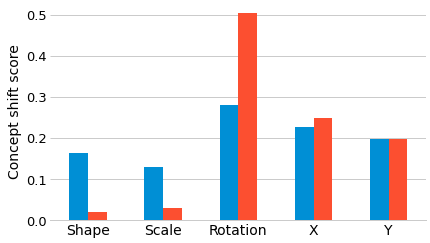} & \includegraphics[width=.33\linewidth]{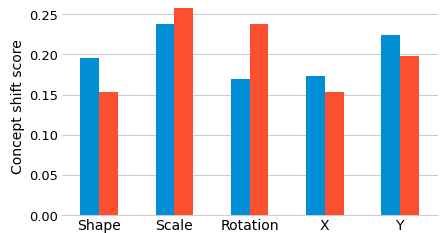} \\
{\scriptsize (d) Gaussian shift  \par} & {\scriptsize (e) Image (shear) shift  \par} & {\scriptsize (f) Image (all) shift  \par} \\[6pt]
\end{tabular}
\caption{Concept shift scores for various knockout, concept, and image shifts. The CSS represents the degree to which a concept experience shifts relative to other concepts. We observe that CBSDs and CBSDh accurately identified and ranked concepts that were affected by shifts.}
\label{fig:appendix-barplot}
\end{figure}

\textbf{Explanations} Figure \ref{fig:appendix-barplot} shows additional explanations for various concept, image, and Gaussian shifts. Similarly to the discussion in Section \ref{sec:result_discussion}, we observe that both CBSD methods accurately identified concepts that were most affected by shift: ``shape" and ``scale" were the most affected concepts when the target data was subjected to a concept (shape, scale) shift; ``x" was the most affected concept when the target data was subjected to a concept (x) shift; ``y" was the most affected concept when the target data was subjected to a concept (y) shift; all concepts were affected when the target data was subjected to all image shifts, namely translation, zoom, shear, and flip. Using the method, we also detected concepts that were affected by image (shear) and Gaussian shifts. We observe that image (shear) shift significantly affected ``rotation" concept, whilst Gaussian shift affected all concepts, with the ``scale" and ``rotation" concepts being the most affected.

\begin{figure}[h!]
\setlength\tabcolsep{0pt}
\begin{tabular}{ccc}
    \multicolumn{3}{c}{\includegraphics[width=.65\linewidth]{figures/legend.png}} \\
  \includegraphics[width=.33\linewidth]{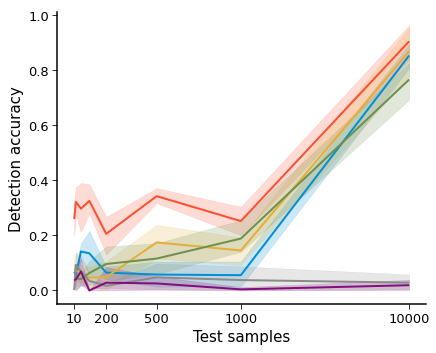} &   \includegraphics[width=.33\linewidth]{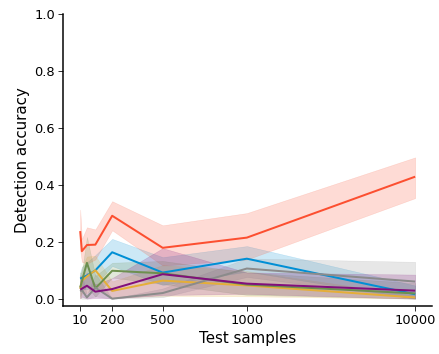} & \includegraphics[width=.33\linewidth]{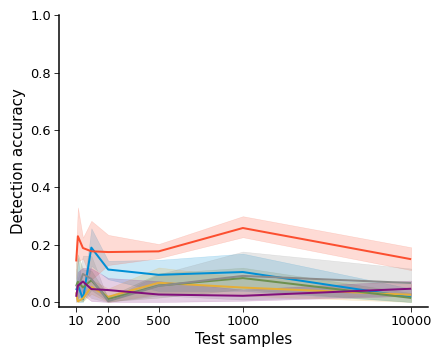} \\
{\scriptsize (a) Knockout shift\par} & {\scriptsize (b) Concept (scale) shift \par} & {\scriptsize (c) Concept (x, y) shift  \par} \\[6pt]
 \includegraphics[width=.33\linewidth]{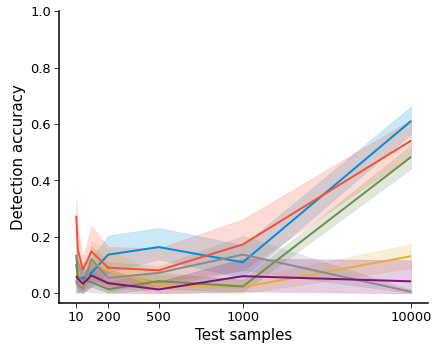} &   \includegraphics[width=.33\linewidth]{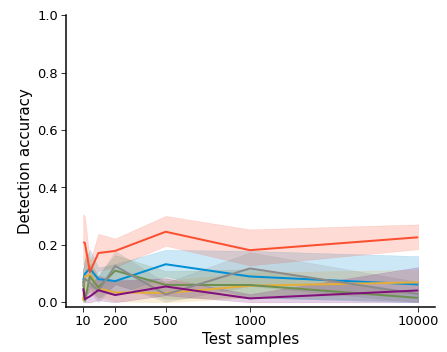} & \includegraphics[width=.33\linewidth]{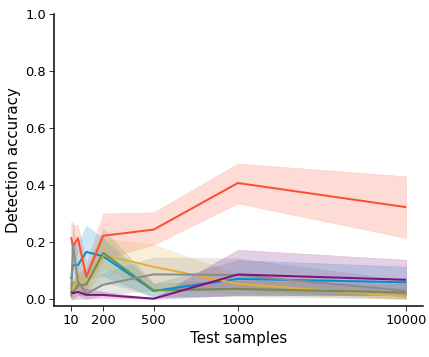} \\
{\scriptsize (d) Concept (shape, scale) shift  \par} & {\scriptsize (e) Concept (x) shift  \par} & {\scriptsize (f) Concept (y) shift  \par} \\[6pt]
\includegraphics[width=.33\linewidth]{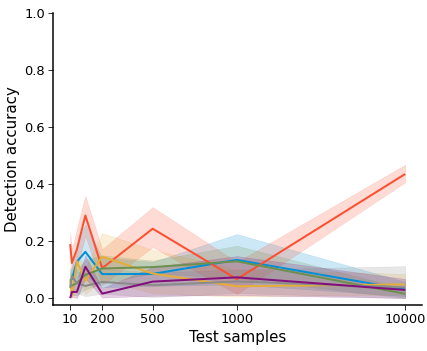} &   \includegraphics[width=.33\linewidth]{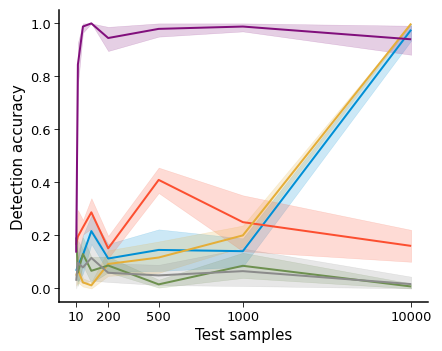} & \includegraphics[width=.33\linewidth]{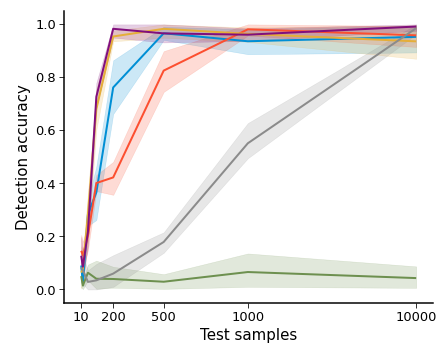} \\
{\scriptsize (g) Concept (scale, x, y) shift \par} & {\scriptsize (h) Gaussian shift  \par} & {\scriptsize (i) Image (all) shift \par} \\[6pt]
\end{tabular}
\caption{Shift detection accuracy for various \textbf{small} knockout, concept, Gaussian, and image shifts. Plots show mean values over 100 random runs with a 95\% confidence interval error bar. Overall, CBSD methods achieved slightly higher detection accuracy than other methods. More significant results are shown in Figures \ref{fig:medium-acc} and \ref{fig:large-acc}.}
\label{fig:small-acc}
\end{figure}

\begin{figure}[h!]
\setlength\tabcolsep{0pt}
\begin{tabular}{ccc}
    \multicolumn{3}{c}{\includegraphics[width=.65\linewidth]{figures/legend.png}} \\
  \includegraphics[width=.33\linewidth]{figures/md-ko.png} &   \includegraphics[width=.33\linewidth]{figures/md-scale.png} & \includegraphics[width=.33\linewidth]{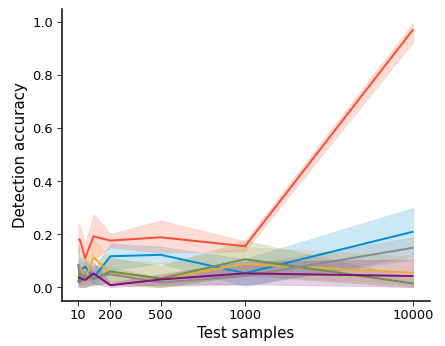} \\
{\scriptsize (a) Knockout shift\par} & {\scriptsize (b) Concept (scale) shift \par} & {\scriptsize (c) Concept (x, y) shift  \par} \\[6pt]
 \includegraphics[width=.33\linewidth]{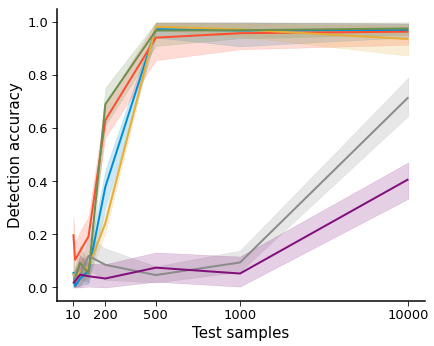} &   \includegraphics[width=.33\linewidth]{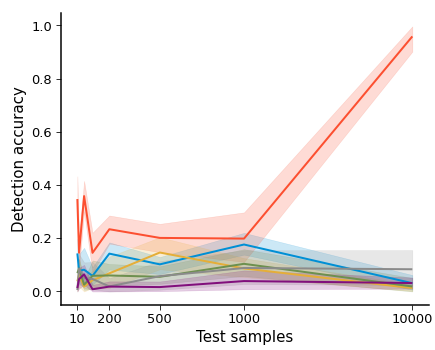} & \includegraphics[width=.33\linewidth]{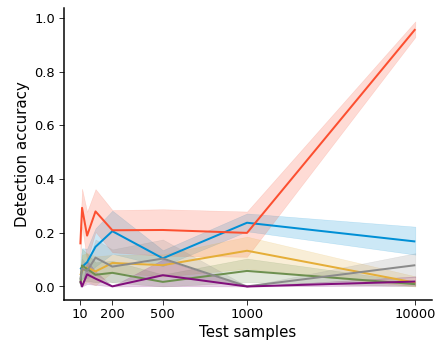} \\
{\scriptsize (d) Concept (shape, scale) shift  \par} & {\scriptsize (e) Concept (x) shift  \par} & {\scriptsize (f) Concept (y) shift  \par} \\[6pt]
\includegraphics[width=.33\linewidth]{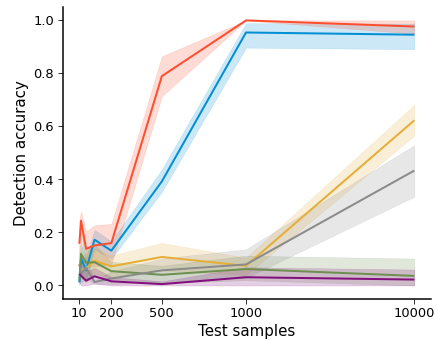} &   \includegraphics[width=.33\linewidth]{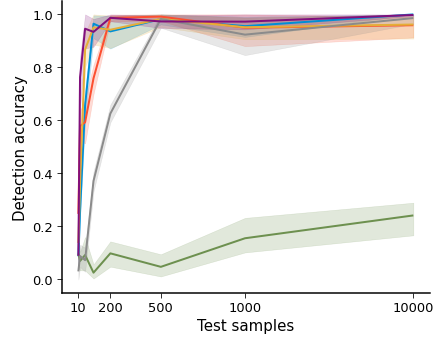} & \includegraphics[width=.33\linewidth]{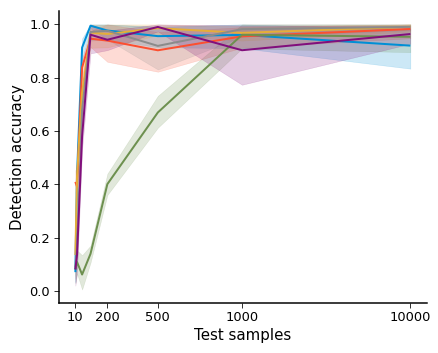} \\
{\scriptsize (g) Concept (scale, x, y) shift \par} & {\scriptsize (h) Gaussian shift  \par} & {\scriptsize (i) Image (all) shift \par} \\[6pt]
\end{tabular}
\caption{Shift detection accuracy for various \textbf{medium} knockout, concept, Gaussian, and image shifts. Plots show mean values over 100 random runs with a 95\% confidence interval error bar.  We observe that BBSD performed similarly to CBSD when shifts directly affect the task labels, shape (see a, d, h, i). In cases where shifts did not directly affect task labels, their performance declined (see b, c, e, f, g).}
\label{fig:medium-acc}
\end{figure}

\begin{figure}[h!]
\setlength\tabcolsep{0pt}
\begin{tabular}{ccc}
    \multicolumn{3}{c}{\includegraphics[width=.65\linewidth]{figures/legend.png}} \\
  \includegraphics[width=.33\linewidth]{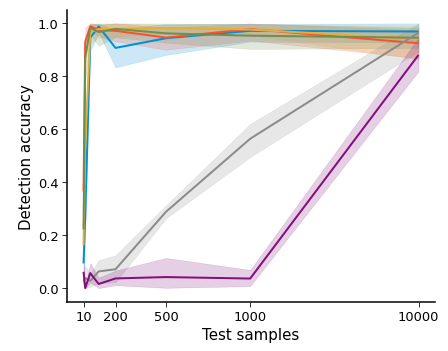} &   \includegraphics[width=.33\linewidth]{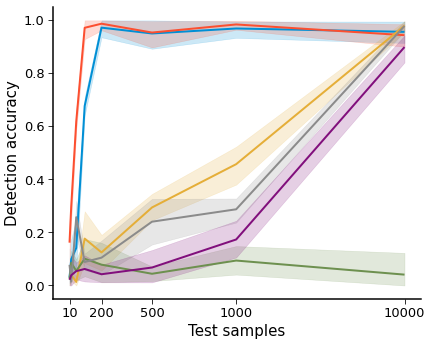} & \includegraphics[width=.33\linewidth]{figures/lg-x_y.png} \\
{\scriptsize (a) Knockout shift\par} & {\scriptsize (b) Concept (scale) shift \par} & {\scriptsize (c) Concept (x, y) shift  \par} \\[6pt]
 \includegraphics[width=.33\linewidth]{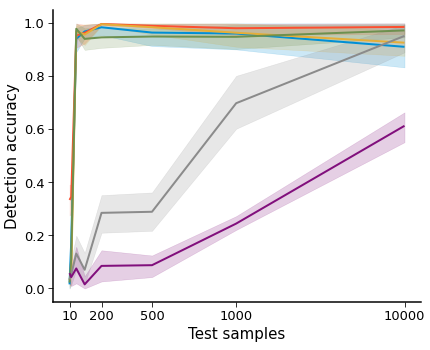} &   \includegraphics[width=.33\linewidth]{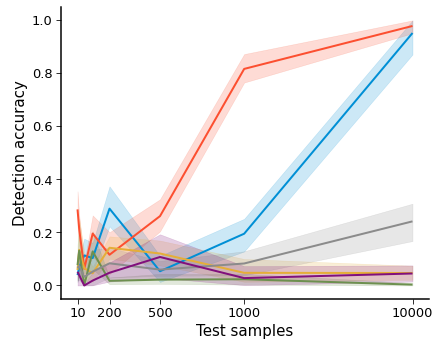} & \includegraphics[width=.33\linewidth]{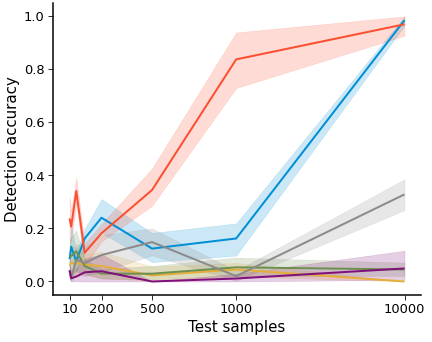} \\
{\scriptsize (d) Concept (shape, scale) shift  \par} & {\scriptsize (e) Concept (x) shift  \par} & {\scriptsize (f) Concept (y) shift  \par} \\[6pt]
\includegraphics[width=.33\linewidth]{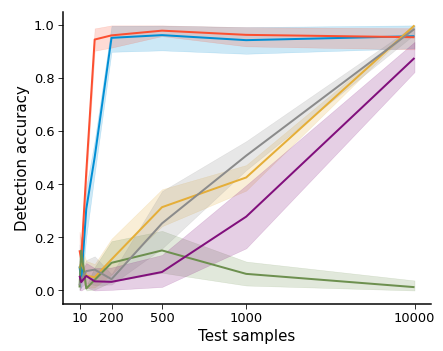} &   \includegraphics[width=.33\linewidth]{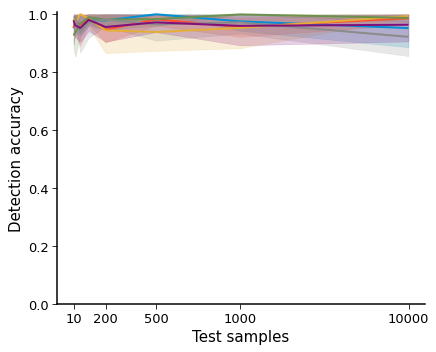} & \includegraphics[width=.33\linewidth]{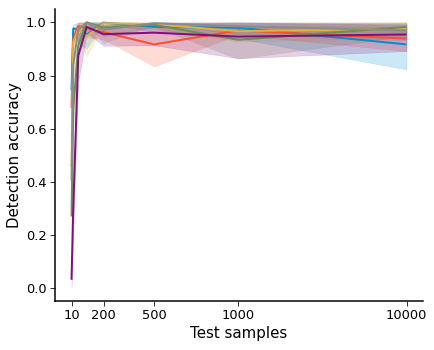} \\
{\scriptsize (g) Concept (scale, x, y) shift \par} & {\scriptsize (h) Gaussian shift  \par} & {\scriptsize (i) Image (all) shift \par} \\[6pt]
\end{tabular}
\caption{Shift detection accuracy for various \textbf{large} knockout, concept, Gaussian, and image shifts. Plots show mean values over 100 random runs with a 95\% confidence interval error bar. We observe that BBSD performed similarly to CBSD when shifts directly affect the task labels, shape (see a, d, h, i). In cases where shifts did not directly affect task labels, their performance declined (see b, c, e, f, g).}
\label{fig:large-acc}
\end{figure}

\textbf{Detection Accuracy} Figures \ref{fig:small-acc}-\ref{fig:large-acc} show shift detection accuracy of all DR methods as a function of test sample sizes for small, medium, and large shift intensities. More accurate results are observed when increasing the number of test samples and when exposing the data with larger shifts. Overall, CBSD methods attained higher shift detection accuracy than BBSD, PCA, and SRP. We observe that CBSD significantly outperform BBSD for detecting shifts that do not directly affect task labels (e.g., concept shifts that do not modify shape). In other cases, BBSD methods performed similarly to CBSD (e.g., for detecting knockout, Gaussian, image, and concept shifts that modify shape).

\begin{figure}[h!]
\begin{center}
\includegraphics[width=1\linewidth]{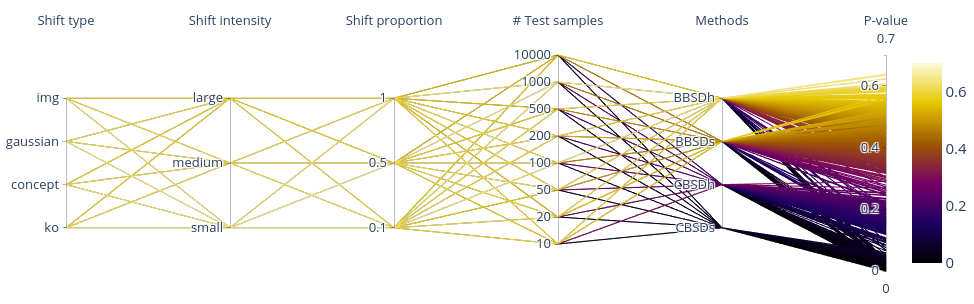}
\end{center}
\caption{Parallel coordinate plot of $p$-values found by BBSDs, BBSDh, CBSDs, CBSDh for various configurations. Overall, we observe that CBSD methods achieved lower $p$-value than BBSD methods, showing more confident shift detection.}
\label{fig:appendix-pval}
\end{figure}

\textbf{$\mathbf{P}$-values} Figure \ref{fig:appendix-pval} visualises $p$-value evolution of different DR methods for various configurations. Overall, CBSDs and CBSDh achieved lower $p$-values compared to other methods for all shift types, intensities, proportions, and test sample sizes. This indicates more confident shift detection, again showing the superiority of CBSD over BBSD; this demonstrates the benefits of using CBM or CME instead of an end-to-end DNN for shift detection.

\FloatBarrier
\subsection{3dshapes}

\begin{figure}[h!]
\setlength\tabcolsep{0pt}
\begin{tabular}{ccc}
    \multicolumn{3}{c}{\includegraphics[width=.25\linewidth]{figures/legend-bar.png}} \\
  \includegraphics[width=.33\linewidth]{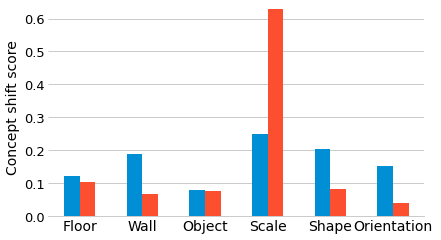} &   \includegraphics[width=.33\linewidth]{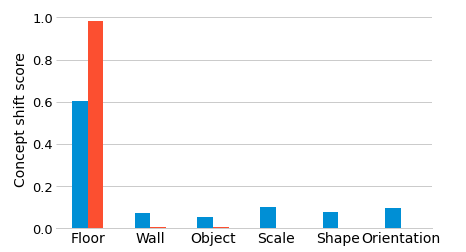} & \includegraphics[width=.33\linewidth]{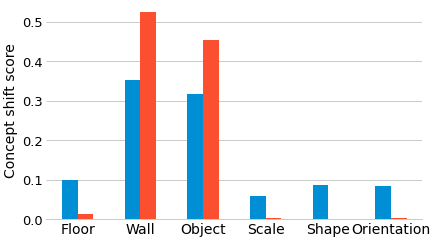} \\
{\scriptsize (a) Knockout shift\par} & {\scriptsize (b) Concept (floor hue) shift \par} & {\scriptsize (c) Concept (wall, object hues) shift  \par} \\[6pt]
 \includegraphics[width=.33\linewidth]{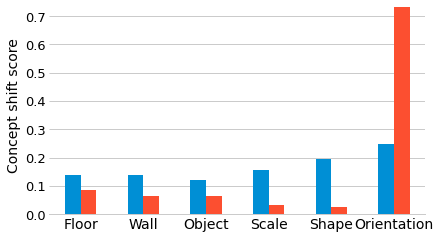} &   \includegraphics[width=.33\linewidth]{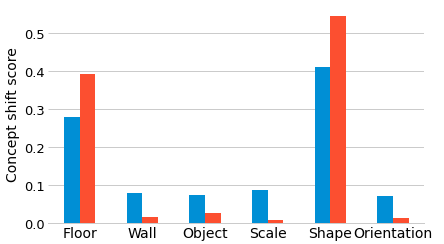} & \includegraphics[width=.33\linewidth]{figures/3d-exp-orientation.png} \\
{\scriptsize (d) Concept (orientation) shift  \par} & {\scriptsize (e) Concept (floor hue, shape) shift  \par} & {\scriptsize (f) Image (rotation) shift  \par} \\[6pt]
\end{tabular}
\caption{Concept shift scores for various knockout, concept, and image shifts. The CSS represents the degree to which a concept experience shifts relative to other concepts. We observe that CBSDs and CBSDh accurately identified and ranked concepts that are affected by shifts.}
\label{fig:3d-barplot}
\end{figure}

\textbf{Explanations} Figure \ref{fig:3d-barplot} shows explanations for various concept, image, and Gaussian shifts for the 3dshapes dataset. Similarly to the discussion in Section \ref{sec:result_discussion} and Appendix \ref{app:dsprites}, we observe that both CBSD methods accurately identified concepts that were most affected by shift: ``floor hue" was the most affected concept when the target data was subjected to a concept (floor hue) shift; ``wall hue" and ``object hue" were the most affected concepts when the target data was subjected to a concept (wall, object hues) shift; ``orientation" was the most affected concept when the target data was subjected to a concept (orientation) shift; ``floor hue" and ``shape" were the most affected concepts when the target data was subjected to a concept (floor hue, shape) shift; ``orientation" was the most affected concept when the target data was subjected to an image (rotation) shift. Interestingly, despite setting the original task to discriminate between all possible scale and shape concept value combinations, we observe that ``scale" was significantly more impacted than the ``shape" concept.

\begin{figure}[h!]
\setlength\tabcolsep{0pt}
\begin{tabular}{ccc}
    \multicolumn{3}{c}{\includegraphics[width=.65\linewidth]{figures/legend.png}} \\
  \includegraphics[width=.33\linewidth]{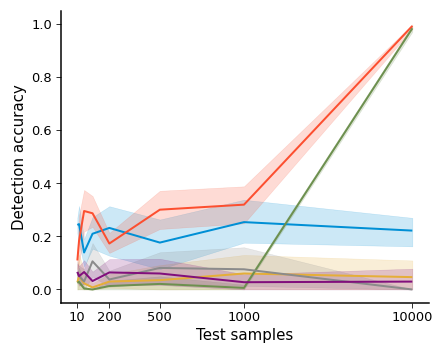} &   \includegraphics[width=.33\linewidth]{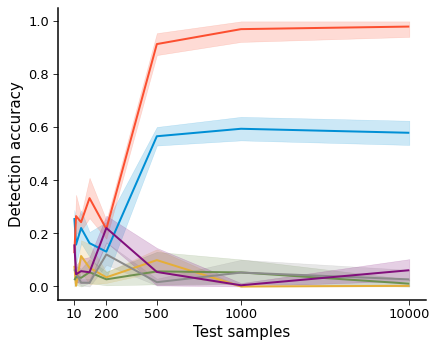} & \includegraphics[width=.33\linewidth]{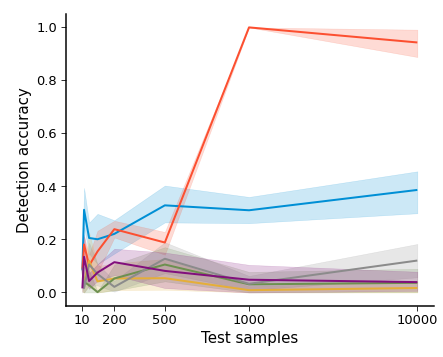} \\
{\scriptsize (a) Knockout shift\par} & {\scriptsize (b) Concept (floor hue) shift \par} & {\scriptsize (c) Concept (wall, object hues) shift  \par} \\[6pt]
 \includegraphics[width=.33\linewidth]{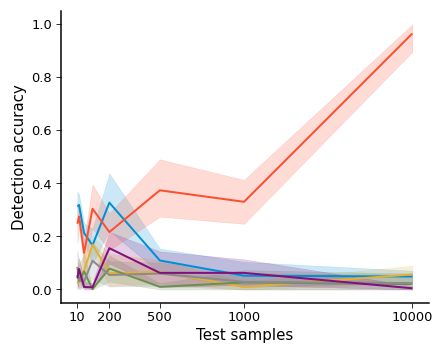} &   \includegraphics[width=.33\linewidth]{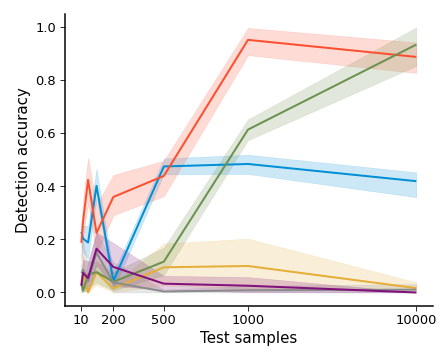} & \includegraphics[width=.33\linewidth]{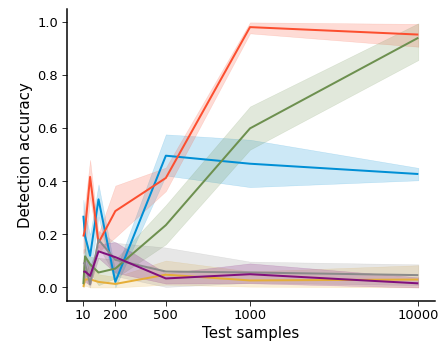} \\
{\scriptsize (d) Concept (orientation) shift  \par} & {\scriptsize (e) Concept (scale) shift  \par} & {\scriptsize (f) Concept (shape) shift  \par} \\[6pt]
\includegraphics[width=.33\linewidth]{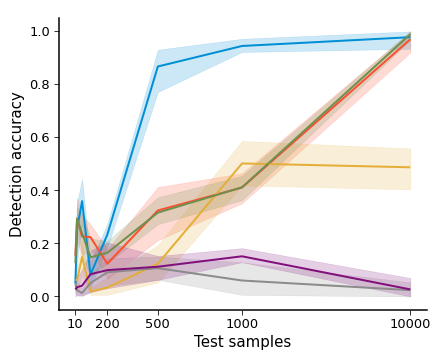} &   \includegraphics[width=.33\linewidth]{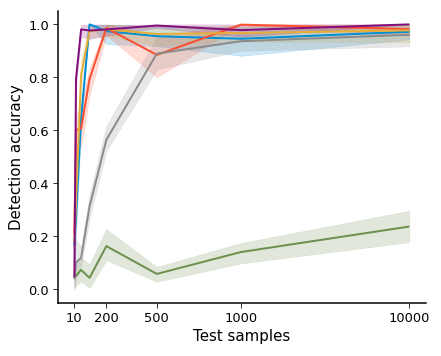} & \includegraphics[width=.33\linewidth]{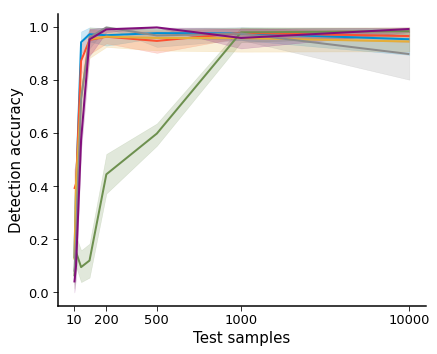} \\
{\scriptsize (g) Concept (shape, floor hue) shift \par} & {\scriptsize (h) Gaussian shift  \par} & {\scriptsize (i) Image (all) shift \par} \\[6pt]
\end{tabular}
\caption{Shift detection accuracy for various \textbf{medium} knockout, concept, Gaussian, and image shifts. Plots show mean values over 100 random runs with a 95\% confidence interval error bar. We observe that BBSD performed similarly to CBSD methods when detecting shifts that are directly related to task labels (a, e, f, g, h, i). In cases, where shifts did not directly affect task labels, their performance declined (b, c, d). Unlike, dSprites, both CBSDh and BBSDh outperformed BBSDs and CBSDs when there is subtle difference between classes (a, e, f).}
\label{fig:3d-medium-acc}
\end{figure}

\begin{figure}[h!]
\setlength\tabcolsep{0pt}
\begin{tabular}{ccc}
    \multicolumn{3}{c}{\includegraphics[width=.65\linewidth]{figures/legend.png}} \\
  \includegraphics[width=.33\linewidth]{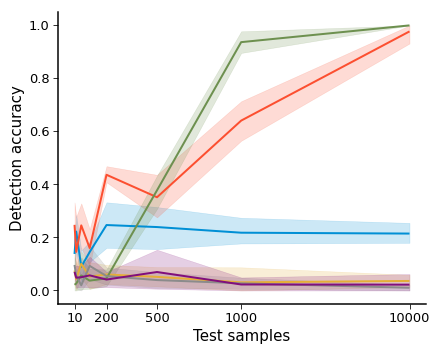} &   \includegraphics[width=.33\linewidth]{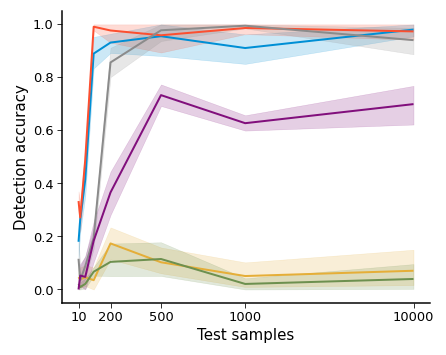} & \includegraphics[width=.33\linewidth]{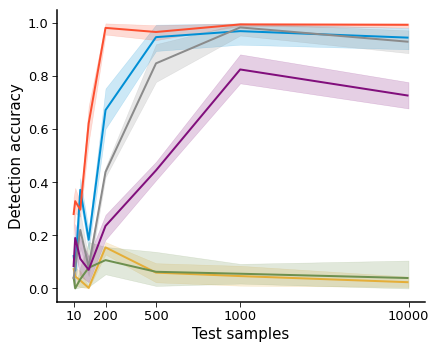} \\
{\scriptsize (a) Knockout shift\par} & {\scriptsize (b) Concept (floor hue) shift \par} & {\scriptsize (c) Concept (wall, object hues) shift  \par} \\[6pt]
 \includegraphics[width=.33\linewidth]{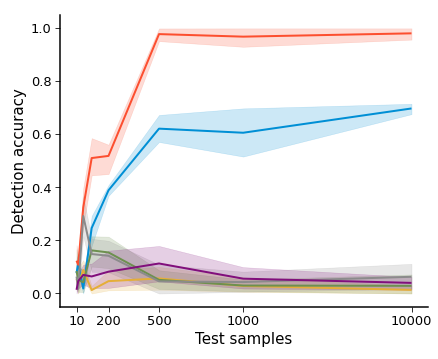} &   \includegraphics[width=.33\linewidth]{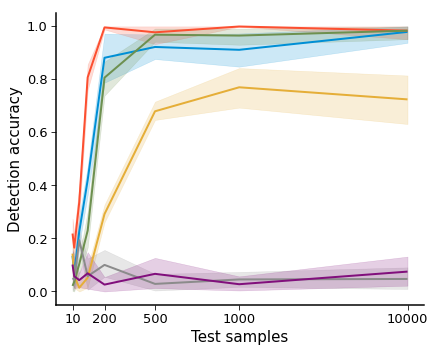} & \includegraphics[width=.33\linewidth]{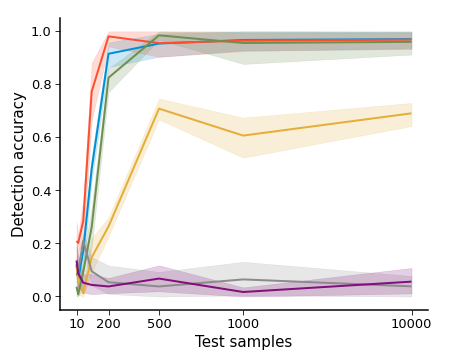} \\
{\scriptsize (d) Concept (orientation) shift  \par} & {\scriptsize (e) Concept (scale) shift  \par} & {\scriptsize (f) Concept (shape) shift  \par} \\[6pt]
\includegraphics[width=.33\linewidth]{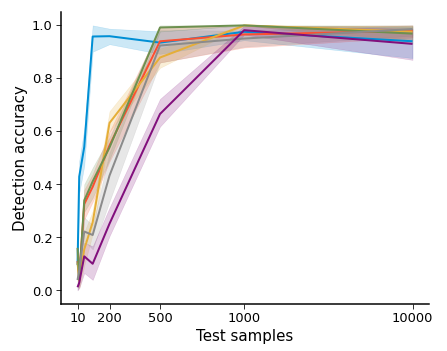} &   \includegraphics[width=.33\linewidth]{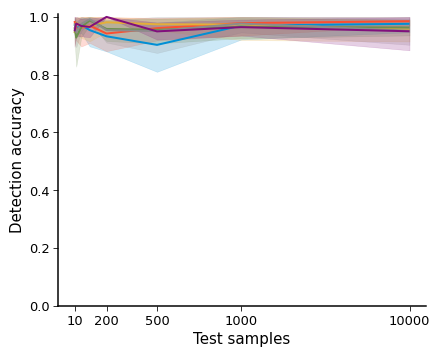} & \includegraphics[width=.33\linewidth]{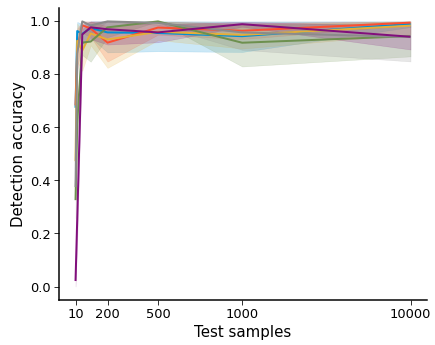} \\
{\scriptsize (g) Concept (shape, floor hue) shift \par} & {\scriptsize (h) Gaussian shift  \par} & {\scriptsize (i) Image (all) shift \par} \\[6pt]
\end{tabular}
\caption{Shift detection accuracy for various \textbf{large} knockout, concept, Gaussian, and image shifts. Plots show mean values over 100 random runs with a 95\% confidence interval error bar. We observe that BBSD performed similarly to CBSD methods when detecting shifts that are directly related to task labels (a, e, f, g, h, i). In cases, where shifts did not directly affect task labels, their performance declined (b, c, d). Unlike, dSprites, both CBSDh and BBSDh outperformed BBSDs and CBSDs when there is subtle difference between classes (a, e, f).}
\label{fig:3d-large-acc}
\end{figure}

\textbf{Detection Accuracy} We arrived at a similar analysis, as discussed in Appendix \ref{app:dsprites}, whereby CBSD methods attained higher shift detection accuracy than BBSD, PCA, and SRP. We observe that CBSD significantly outperforms BBSD for detecting shifts that do not directly affect task labels. In other cases, BBSD methods performed similarly to CBSD. Unlike previous sections, hard-thresholded prediction (CBSDh and BBSDh) performed significantly better than the softmax variants (CBSDs and BBSDs), visualises in Figures \ref{fig:3d-medium-acc} and \ref{fig:3d-large-acc} (a, e, f). This is due to the subtle difference between classes, resulting in boundary softmax values, making it harder for the statistical tests to conclude difference in distributions. Conversely, the hard-thresholded variant is unaffected, as they discretise decisions.

\textbf{$\mathbf{P}$-values} Figure \ref{fig:appendix-pval} visualises $p$-value evolution of different DR methods for various configurations. We observe same results as Appendix \ref{app:dsprites}, where CBSDs and CBSDh achieved lower $p$-value compared to other methods for all shift types, intensities, proportions, and number of test samples, indicating more confident shift detection.

\begin{figure}[h!]
\begin{center}
\includegraphics[width=1\linewidth]{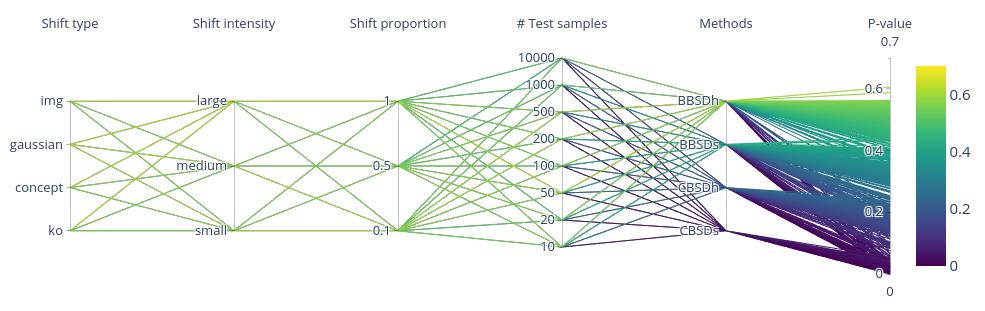}
\end{center}
\caption{Parallel coordinate plot of $p$-values found by BBSDs, BBSDh, CBSDs, CBSDh for various configurations. Similarly to Figure~\ref{fig:appendix-pval}, we observe that CBSD methods achieved lower $p$-value than BBSD methods, signifying more confident shift detection.}
\label{fig:appendix-pval-2}
\end{figure}

\begin{figure}[h!]
\setlength\tabcolsep{0pt}
\begin{tabular}{ccc}
    \multicolumn{3}{c}{\includegraphics[width=.65\linewidth]{figures/legend.png}} \\
  \includegraphics[width=.33\linewidth]{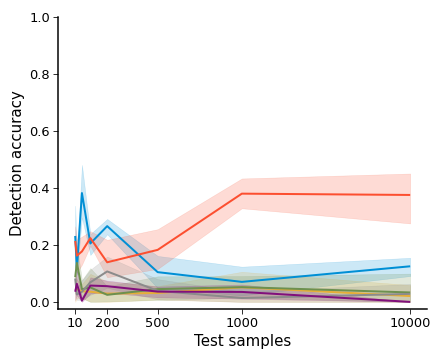} &   \includegraphics[width=.33\linewidth]{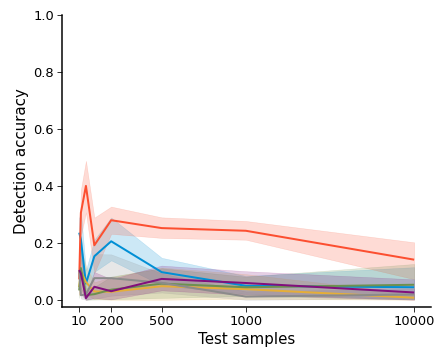} & \includegraphics[width=.33\linewidth]{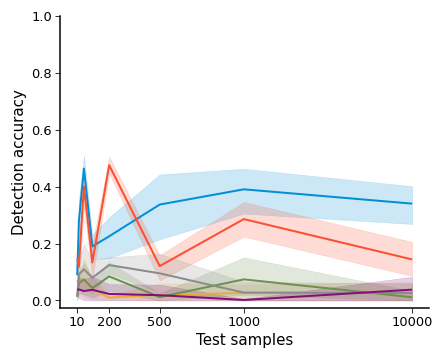} \\
{\scriptsize (a) Knockout shift\par} & {\scriptsize (b) Concept (floor hue) shift \par} & {\scriptsize (c) Concept (wall, object hues) shift  \par} \\[6pt]
 \includegraphics[width=.33\linewidth]{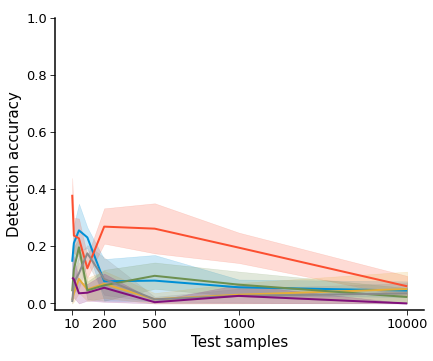} &   \includegraphics[width=.33\linewidth]{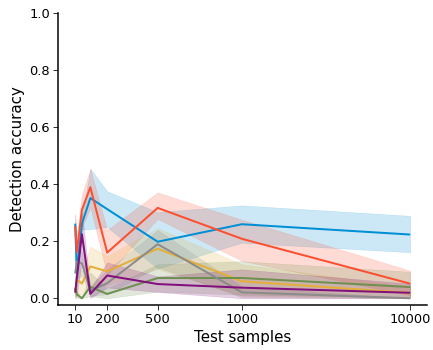} & \includegraphics[width=.33\linewidth]{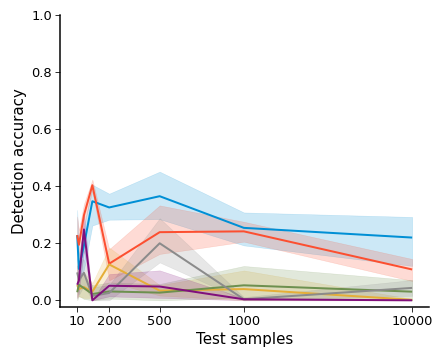} \\
{\scriptsize (d) Concept (orientation) shift  \par} & {\scriptsize (e) Concept (scale) shift  \par} & {\scriptsize (f) Concept (shape) shift  \par} \\[6pt]
\includegraphics[width=.33\linewidth]{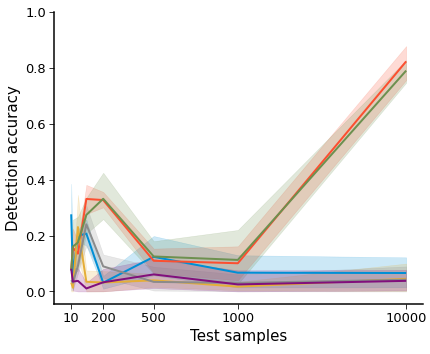} &   \includegraphics[width=.33\linewidth]{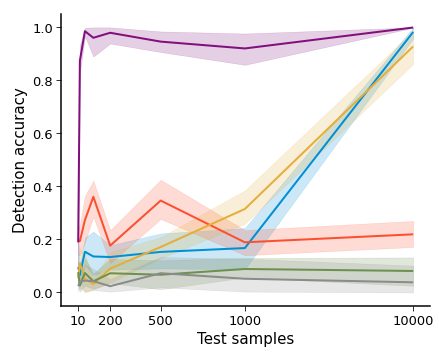} & \includegraphics[width=.33\linewidth]{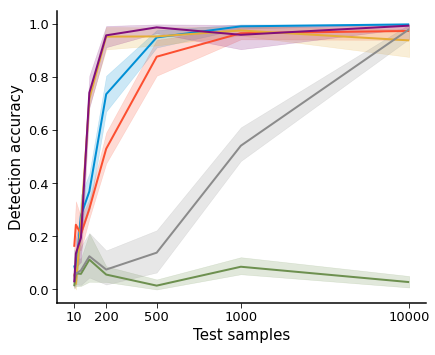} \\
{\scriptsize (g) Concept (shape, floor hue) shift \par} & {\scriptsize (h) Gaussian shift  \par} & {\scriptsize (i) Image (all) shift \par} \\[6pt]
\end{tabular}
\caption{Shift detection accuracy for various \textbf{small} knockout, concept, Gaussian, and image shifts. Plots show mean values over 100 random runs with a 95\% confidence interval error bar. Overall, CBSD methods achieved slightly higher detection accuracy than other methods. More significant results are shown in Figures \ref{fig:3d-medium-acc} and \ref{fig:3d-large-acc}.}
\label{fig:3d-small-acc}
\end{figure}

\end{document}